\pdfoutput=1

\documentclass[11pt]{article}
\usepackage[]{EMNLP2022}

\usepackage{times}
\usepackage{latexsym}

\usepackage[T1]{fontenc}

\usepackage[utf8]{inputenc}

\usepackage{microtype}

\usepackage{inconsolata}
\usepackage{xcolor}
\usepackage{adjustbox}
\usepackage{diagbox}
\usepackage{amsmath,amssymb}
\usepackage[inline]{enumitem}
\usepackage{subcaption}
\usepackage{hypcap}
\usepackage{arydshln}
\usepackage{pifont}
\usepackage{multirow}

\newcommand{\rbert}{\texttt{Vanilla}}
\newcommand{\bert}{\texttt{Certainty}}
\newcommand{\rdt}{\texttt{DT}}
\newcommand{\dt}{\texttt{Cert-DT}}
\newcommand{\rdann}{\texttt{DANN}}
\newcommand{\dann}{\texttt{Cert-DANN}}
\newcommand{\dropout}{\texttt{MC-Dropout}}
\newcommand{\smoothing}{\texttt{Smoothing}}
\newcommand{\select}{\texttt{Selection}}
\newcommand{\tapad}{\texttt{Cert-TAPAD}}
\newcommand{\rtapad}{\texttt{TAPAD}}

\newcommand{\cmark}{\ding{51}}
\newcommand{\xmark}{\ding{55}}

\title{Domain Adaptation from Scratch}

\author{Eyal Ben-David \textsuperscript{\ 1}, Yftah Ziser\textsuperscript{2}, Roi Reichart\textsuperscript{1} \\
  \textsuperscript{1}Faculty of Industrial Engineering and Management, Technion, IIT \\
  \textsuperscript{2}School of Informatics, University of Edinburgh \\
  \texttt{\{eyalbd12@campus.|roiri@\}technion.ac.il} \\
  \texttt{yftah.ziser@ed.ac.uk} \\}

\begin{document}
\maketitle

\begin{abstract}
Natural language processing (NLP) algorithms are rapidly improving but often struggle when applied to out-of-distribution examples. A prominent approach to mitigate the domain gap is domain adaptation, where a model trained on a source domain is adapted to a new target domain. We present a new learning setup, ``domain adaptation from scratch'', which we believe to be crucial for extending the reach of NLP to sensitive domains in a privacy-preserving manner.
In this setup, we aim to efficiently annotate data from a set of source domains such that the trained model performs well on a sensitive target domain from which data is unavailable for annotation. Our study compares several approaches for this challenging setup, ranging from data selection and domain adaptation algorithms to active learning paradigms, on two NLP tasks: sentiment analysis and Named Entity Recognition. Our results suggest that using the abovementioned approaches eases the domain gap, and combining them further improves the results.
\footnote{Our code and data are available at \url{https://github.com/eyalbd2/ScratchDA}.}

\end{abstract}

\section{Introduction}
\label{sec:intro}

Natural Language Processing (NLP) algorithms have made impressive progress in recent years, achieving great success in various tasks and applications. However, these algorithms may require substantial labeled data to reach their full potential. Such data is often too costly and labor-intensive to obtain. Furthermore, modern systems are required to perform well on data from many domains, while for most tasks, labeled data is available for only a few. Consequently, NLP algorithms still struggle when applied to test examples from unseen domains \citep{dilbert, pada, docogen}. 

A common strategy for annotating data is through crowdsourcing platforms, e.g., Amazon Mechanical Turk, where data can be annotated on a large-scale, as one can harness a large workforce at affordable prices. However, crowdsourcing is not an option for sensitive data (e.g., in medical and military applications). For instance, virtual assistants such as Alexa, Siri, and Google Home serve millions of users. Annotating data from virtual assistants' recorded conversations would be optimal input for training applicable NLP models for their purposes. However, crowdsourcing is not a valid option for such sensitive data as it does not preserve users' privacy. Hiring annotators internally is expensive and often hard to scale efficiently.  

This paper examines different approaches to annotating and using data from non-sensitive sources (NSS) when \emph{target domain data can not be annotated}. More precisely, we experiment with several methods (\S\ref{sec:methods}) to train a model for a sensitive target domain under a given annotation budget of examples from NNS. Various methods aim to ease the burden of data annotation. 
For example, domain adaptation (DA) algorithms are trained on source domain training data to be effectively applied to other target domains. 
Several setups have been explored in the DA literature,  differing in the availability of target domain supervision. 
This paper focuses on unsupervised DA (UDA), i.e., where no labeled data is available from the target domain. Still, unlabeled data can be freely obtained from both source and target domains. Unlike previous works on UDA \cite{ziser-pblm, perl}, trying to train the best model for a given labeled source data, we are interested in efficiently annotating the source data. Hence, we perform domain adaptation from scratch. 



Another prominent approach for coping with limited annotation resources is  Active Learning (AL, \citep{DBLP:journals/jair/CohnGJ96}). AL algorithms automatically choose a subset of unlabeled samples from a massive unlabeled corpus, maximizing the value of the labeling process. Such algorithms have shown effectiveness for deep neural network (DNN) modeling across various NLP tasks \citep{DBLP:conf/acl/DuongAEPCJ18, DBLP:conf/conll/PerisC18, DBLP:conf/emnlp/Ein-DorHGSDCDAK20}. 
The typical assumption in the AL literature is that AL algorithms choose the most valuable subset to annotate under the assumption of being tested on data from the same distribution. This paper examines their usefulness in improving model performance when applied to unknown target domains. 

As DA and AL are highly popular approaches to ease the problem of insufficient annotation budget, combining the two is natural \cite{rai2010domain, saha2011active}. 
Most works in this intersection share a similar setup, in which labeled data from a source domain facilitates the annotation process in the target domain \citep{li2013active, wu2017active}.
\citet{xiao2013online} consider 
data annotation to be cheaper in the source domain compared to the target domain.
We consider this work the most similar to ours since it recognizes the cost of annotating samples from the source domain for improving in the target domain. However, annotating samples from the target domain is unavailable in our setup.

We propose and evaluate various methods to tackle the challenging ``DA from scratch'' setup, focusing on two tasks: Binary sentiment classification and named entity recognition. 
Going beyond existing methods, we propose several methods for integrating domain knowledge into AL systems and demonstrate encouraging results:  Combining the two is better than applying them separately. We hope to inspire future work in this direction.

\section{Problem Definition}
\label{sec:problem}
The goal of a DA algorithms is to learn a model $m$ that generalizes well from a set of source domains  $\{\mathcal{D}_{S_i}\}_{i=1}^{K}$ to a  target domain $\mathcal{D}_T$, for a given task. 
We assume that unlabeled data from the source and target domains are available. 
We denote the corresponding unlabeled sets with $\{\mathcal{U}_{S_i}\}_{i=1}^{K}$  and $ \mathcal{U}_T $ respectively. Initially, no labeled data exists, and since data from the target domain is sensitive, we cannot annotate it. Thus, we consider an approach for automatically choosing  $L$ source domain examples to annotate in each iteration. We choose examples that would best benefit a model applied to test data that originates from the target domain.  

\begin{table}
\small
\centering
\begin{adjustbox}{width=0.35\textwidth}

\begin{tabular}{| l | l | c | c | c | c | c|}
\hline
  & & \multicolumn{4}{c|}{Method} \\
   Section & Name  & \multicolumn{1}{c}{\rotatebox[origin=c]{90}{\tiny{DA}}} & \multicolumn{1}{c}{\rotatebox[origin=c]{90}{\tiny{AL}}}  & \multicolumn{1}{c}{\rotatebox[origin=c]{90}{\tiny{NIS}}} &\rotatebox[origin=c]{90}{\tiny{DS}} \\
\hline

\multirow{1}{*}{\S\ref{sec:naive}} &\rbert & \xmark  &  \xmark  & \xmark & \xmark \\
\hline

\multirow{3}{*}{\S\ref{sec:basic-al}} &\bert & \xmark &  \cmark   & \xmark & \xmark \\
&\dropout & \xmark  & \cmark & \xmark &  \xmark \\
&\smoothing & \xmark &   \cmark  & \xmark & \xmark \\
\hline

\multirow{2}{*}{\S\ref{sec:da-no-al}} &\rdt & \cmark &   \xmark  & \xmark & \xmark \\
&\rdann & \cmark &  \xmark  & \xmark & \xmark \\
\hline

\multirow{2}{*}{\S\ref{sec:iterative-certainty}} &\dt & \cmark &  \cmark   & \xmark & \xmark \\
&\dann & \cmark &   \cmark  & \xmark & \xmark \\
\hdashline

\multirow{3}{*}{\S\ref{sec:domain-similarity}} &\select & \cmark &   \cmark  & \cmark & \xmark \\
&\rtapad & \cmark &  \xmark  & \xmark & \cmark \\
&\tapad & \cmark &  \cmark   & \xmark & \cmark \\

\hline
\end{tabular}
\end{adjustbox}
\caption{Our methods. We specify which use domain adaptation (DA), active learning (AL), non-iterative example selection (NIS), and domain selection (DS).}
\label{tab:methods}
\end{table}

\section{Methods}
\label{sec:methods}

Table~\ref{tab:methods} outlines our methods, which are divided into four branches: (1) Supervised learning with naive labeling (\S\ref{sec:naive}); (2) active learning methods (\S\ref{sec:basic-al}); (3) domain adaptation approaches (\S\ref{sec:da-no-al}); and (4) methods that combine domain adaptation with  AL (\S\ref{sec:DA-and-AL}).\footnote{There are two folk wisdom definitions of AL. The first clings to sample selection while the second advocates interactivity. For simplicity, we follow the first definition.} 
We implement all our algorithms with a pretrained BERT model \citep{bert}.

\subsection{Supervised Learning with Naive Labeling}
\label{sec:naive}
We first apply a straightforward approach. 
Particularly, we randomly choose which examples to annotate and finetune an out-of-the-box BERT on the labeled data. We name this baseline \textbf{\rbert}.

\subsection{Active Learning Approaches}
\label{sec:basic-al}
We implement a set of algorithms that perform example annotation according to the model's certainty. We test the following models: (a) \textbf{\bert}, iteratively choosing examples according to the (in)confidence of a BERT trained on the available annotated examples (so far); (b) \textbf{\dropout} (\textit{monte-carlo dropout}, \citep{DBLP:conf/icml/GalG16}), applying many dropout instances of the example to get a more robust certainty score of the model; and (c) \textbf{ \smoothing} (\textit{label smoothing}), a variant of \bert{}  but when the underlying NLP model is trained with a label smoothing objective.

\begin{figure}

\begin{minipage}{.48\textwidth}
\centering
\includegraphics[scale=.33]{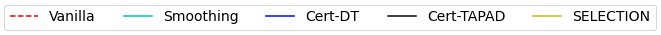}
\end{minipage}

\begin{minipage}{.24\textwidth}
\centering
\subfloat[\textbf{DVD}]{\label{fig:dvd}\includegraphics[scale=.18]{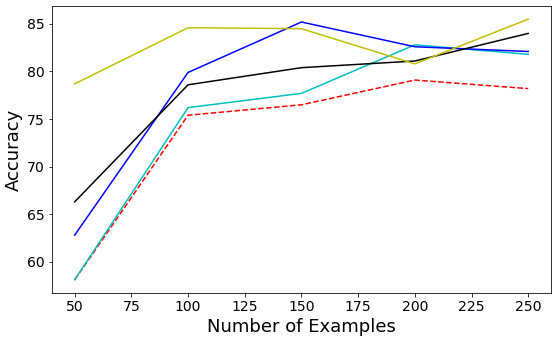}}
\end{minipage}%
\begin{minipage}{.24\textwidth}
\centering
\subfloat[\textbf{Electronics}]{\label{fig:electronics}\includegraphics[scale=.18]{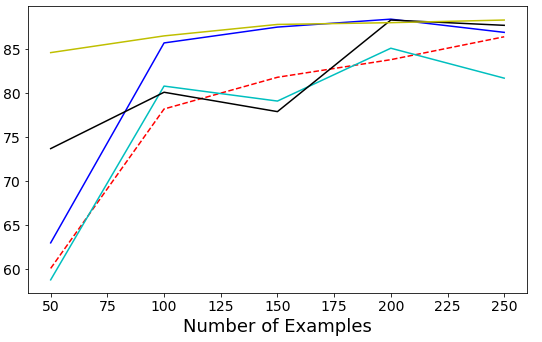}}
\end{minipage}

\begin{minipage}{.24\textwidth}
\centering
\subfloat[\textbf{BC}]{\label{fig:bc}\includegraphics[scale=.18]{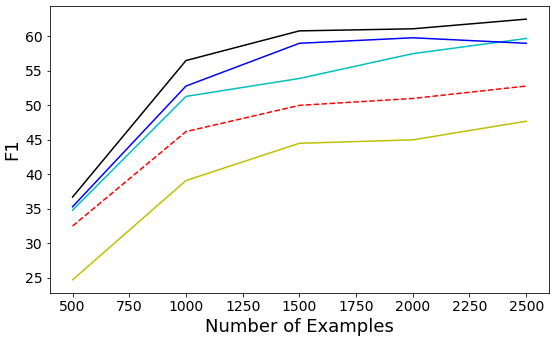}}
\end{minipage}
\begin{minipage}{.24\textwidth}
\centering
\subfloat[\textbf{NW}]{\label{fig:nw}\includegraphics[scale=.18]{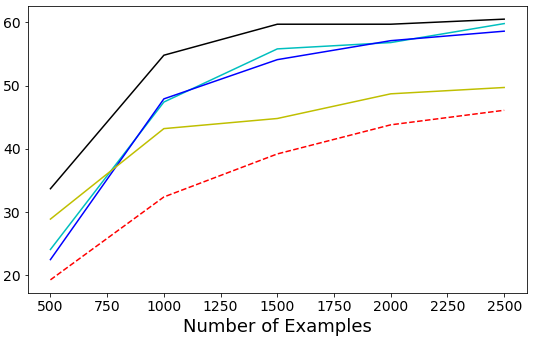}}
\end{minipage} \par\medskip
\caption{Sentiment classification (a-b) and NER (c-d) results with a breakdown to active learning iteration.}
\label{fig:sa-ner-main}
\end{figure}

\subsection{Domain Adaptation}
\label{sec:da-no-al}
UDA algorithms require unlabeled data from the source and target domains and labeled data only from the source domain, to be applied to test data from the target domain. 
We implement two popular UDA algorithms: (1) \textbf{\rdt} (\textit{domain-tuned} \citep{DBLP:conf/emnlp/HanE19}), which is first finetuned on the available unlabeled data (from the source and target domains) by performing the language modeling task.
Then, it is further finetuned on task labeled data from the source domains; (2) \textbf{\rdann}  (\textit{domain adversarial neural networks} \citep{dann}), a model that integrates BERT with an adversarial domain classifier to learn domain invariant representations. \rdann{} performs multi-task training, combining the downstream task and the adversarial objectives. 
For both methods, we randomly select the source domain examples for annotation.
\subsection{Combining DA and AL}
\label{sec:DA-and-AL}
We next introduce novel approaches for incorporating DA methods within AL algorithm. We distinguish between iterative and non-iterative example selection  methods (IS and NIS). 
Finally, we present a domain-selection (DS) method, choosing a single source domain to represent the annotation pool.

\subsubsection{Iterative Certainty-based Algorithms}
\label{sec:iterative-certainty}
We present \textbf{\dt{}} and \textbf{\dann}, UDA models corresponding to the ones described in \S\ref{sec:da-no-al} and combined with \bert. Unlike \rdt{} and \rdann, where the source domain annotated examples are randomly selected, these models train on labeled data iteratively selected by their confidence scores.


\subsubsection{Leveraging Domain Similarities}
\label{sec:domain-similarity}

Utilizing domain-similarity metrics 
to select the most adaptable examples for annotation,  lies at the intersection
of AL and DA. $A-$distance \citep{DBLP:conf/vldb/KiferBG04} is a standard metric for measuring similarity between domains. \citet{ben2007analysis} interpret the \textit{proxy $A-$distance (PAD)} between two domains as the loss a linear classifier achieves when predicting the origin domain of samples obtained from both domains. As PAD is task-independent, it can not harness any information about the task at hand. To address this problem, \citet{elsahar2019annotate} introduced the  \textit{task-adapted Proxy $A-$distance} (TAPAD)\footnote{\citet{elsahar2019annotate}  named this measure PAD*.} measure, which trains the domain-discrimination model on representations obtained from a model which is finetuned for a downstream task. TAPAD has been demonstrated to effectively predict the performance degradation for cross-domain setups.

\paragraph{Task-adapted Proxy $A-$distance (TAPAD) }
We propose a novel integration of TAPAD as part of an AL mechanism. After randomly sampling the initial annotation seed, we train a model for the downstream task. Then, we tune the model for discriminating between source and target examples, using unlabeled data from the source and target domains. We then measure the TAPAD score between each of the source domains and the target domain.  Finally, we constrain the annotation pool to contain data only from the most similar source domain for the rest of the annotation process. 
We implement two methods for selecting the data for annotation from the most similar domain: either randomly, with \textbf{\rtapad}, or using an uncertainty measure, with \textbf{\tapad}.

\paragraph{Non-iterative Example Selection}
We implement a domain discriminator trained on unlabeled data from the source and target domains. The predictor is then used to provide a score $\mathcal{S}(x)$ for each example ($x$, from the source domains), where  $\mathcal{S}(x) = \frac{1}{M(x)}$ and $M(x)$ is the predictor's output. The model scores range from $0$ to $1$, higher scores mean the model is more confident that the example
originates from the corresponding source domain.
We annotate the top $L$ ranked unlabeled examples. 
Since the scores are not affected by new annotations, we do not need to re-rank the examples iteratively. We name this algorithm \textbf{\select}.

\begin{table*}

\centering
\begin{adjustbox}{width=0.87\textwidth}

\begin{tabular}{ | l || c | c | c | c | c || c || c | c | c | c | c | c || c |}
\hline
& \multicolumn{6}{c||}{\textbf{Sentiment}} & \multicolumn{7}{c|}{\textbf{NER}} \\ 
\hline
  & \textbf{A}  & \textbf{B} & \textbf{D} & \textbf{E} & \textbf{K} & \textbf{AVG}  & \textbf{BC}  & \textbf{BN} & \textbf{MZ} & \textbf{NW} & \textbf{TC} & \textbf{WB} & \textbf{AVG}  \\
\hline

\textbf{\rbert} & $ 77.0 $ &  $ 75.0 $ & $ 72.3 $ & $ 76.0 $ & $ 78.1 $ & $ 75.4 $ & $ 46.5 $ &  $ 48.5 $ & $ 42.1 $ & $ 36.2 $ & $ 32.5 $ & $ 30.7 $ & $ 39.4 $  \\
\hline
\textbf{\bert}     & $ 75.8 $ &  $ 75.7 $ & $ 73.7 $ & $ 76.2 $ & $ 76.7 $ & $ 75.3 $ & $ 50.9 $ &  $ 55.1 $ & $ 45.2 $ & $ 47.1 $ & $ 32.8 $ & $ 34.1 $ & $ 44.2 $  \\
\textbf{\smoothing}     & $ 75.1 $ &  $ 74.7 $ & $ 75.3 $ & $ 77.1 $ & $ 77.6 $ & $ 76.0 $ & $ 51.4 $ &  $ 55.7 $ & $ 47.0 $ & $ 48.8 $ & $ 33.8 $ & $ 33.7 $ & $ 45.1 $  \\
\textbf{\dropout}     & $ 76.7 $ &  $ 72.9 $ & $ 72.7 $ & $ 79.0 $ & $ 79.8 $ & $ 76.2 $ & $ 51.1 $ &  $ 54.8 $ & $ 43.7 $ & $ 46.7 $ & $ 32.5 $ & $ 33.6 $ & $ 43.7 $  \\
\hline
\textbf{\rdt}   & $ \textbf{78.0} $ &  $ 78.1 $ & $ 76.2 $ & $ 79.5 $ & $ 82.1 $ & $ 78.6 $ & $ 47.7 $ &  $ 51.0 $ & $ 43.0 $ & $ 38.4 $ & $ 33.8 $ & $ 30.8 $ & $ 40.8 $  \\
\textbf{\rdann} & $ \textbf{78.0} $ &  $ 75.7 $ & $ 77.0 $ & $ 79.5 $ & $ 79.4 $ & $ 77.9 $ & $ 45.9 $ &  $ 46.5 $ & $ 40.4 $ & $ 35.7 $ & $ 32.0 $ & $ 29.1 $ & $ 38.2 $  \\
\hline
\textbf{\dt}       & $ \textbf{78.0} $ &  $ 77.5 $ & $ 77.6 $ & $ 81.2 $ & $ 81.9 $ & $ 79.0 $ & $ 53.2 $ &  $ \textbf{56.3} $ & $ 46.6 $ & $ 48.0 $ & $ 34.4 $ & $ \textbf{34.1} $ & $ 45.4 $  \\
\textbf{\dann}     & $ 75.0 $ &  $ 76.8 $ & $ 77.2 $ & $ 77.8 $ & $ 80.8 $ & $ 77.5 $ & $ 50.9 $ &  $ 54.6 $ & $ 45.9 $ & $ 47.3 $ & $ 32.3 $ & $ 33.2 $ & $ 44.0 $  \\
\hline
\textbf{\select}       & $ 76.9 $ &  $ \textbf{79.7} $ & $ \textbf{82.8} $ & $ \textbf{87.0} $ & $ \textbf{87.0} $ & $ \textbf{82.7} $ & $ 40.2 $ &  $ 48.2 $ & $ 40.9 $ & $ 43.1 $ & $ 12.8 $ & $ 28.4 $ & $ 35.6 $  \\
\textbf{\rtapad}       & $ 77.6 $ &  $ 76.5 $ & $76.2 $ & $ 82.2 $ & $ 80.9 $ & $ 78.7 $ & $ 50.2 $ &  $ 50.3 $ & $ 45.9 $ & $ 48.7 $ & $ 31.2 $ & $ 26.3 $ & $ 42.1 $  \\
\textbf{\tapad}       & $ 76.9 $ &  $ 74.5 $ & $ 78.1 $ & $ 81.5 $ & $ 80.6 $ & $ 78.3 $ & $ \textbf{55.5} $ &  $ 54.7 $ & $ \textbf{49.4} $ & $ \textbf{53.7} $ & $ 33.2 $ & $ 30.4 $ & $ \textbf{46.2} $  \\
\hline

\hline
\end{tabular}
\end{adjustbox}
\caption{AUC - SA (up to 250 examples) and NER (up to 2500 examples).}
\label{tab:main-results}
\end{table*}

\section{Experimental Setup}
We focus on two common NLP tasks, (binary) sentiment analysis (SA) and named entity recognition (NER). For SA, 
following a large body of prior work \citep{ziser-reichart-2018-deep, ziser19, perl}, 
we include 
four product review domains \citep{blitzer2007biographies} - Books (B), DVDs (D), Electronics (E) and Kitchen appliances (K); and an additional airline review dataset (A,  \cite{Nguyen2015airline}).
For NER, we experiment with the OntoNotes 5.0 dataset \citep{ontonotes}, including six different domains: Broadcast Conversations (BC), Broadcast News (BN), Magazine (MZ), Newswire (NW), Telephone Conversation (TC), and Web (WB). We report the statistics of our experimental setups in \S\ref{sec:implementation}.\footnote{URLs of the datasets, code, implementation details, and hyperparameter configurations are described in \S~\ref{sec:implementation}.
}

We choose a single target domain that is held during training and used only for testing. Accordingly, all other domains form our source domains and are used for training. We repeat the experiments in each task to use each domain as the target domain.
We start each IS algorithm by randomly sampling $M$ examples and annotating them ($M=50$ and $M=250$ for SA and NER, respectively). Afterward, we choose $M$ examples to annotate in each iteration according to the method in use. 
We report accuracy scores for sentiment classification\footnote{We use balanced test sets for sentiment analysis.} and macro-F1 scores for NER. All reported scores are averaged across five different seeds. To conclude the accumulated results across different iterations, we report the AUC score for each model.

\section{Results and Analysis}
\label{sec:results}

Table~\ref{tab:main-results} presents our main results. 
In Figure~\ref{fig:sa-ner-main}, we present the results of two representative setups for each task, with a breakdown to iterations.
As our results indicate, methods that integrate DA and AL, which is a novel contribution of this paper, performs best on both tasks.

For both tasks, Figure~\ref{fig:sa-ner-main} demonstrates the effectiveness our methods compared to \rbert{} (AL, DA, and their combination) across (almost) all active learning iterations.
In NER, the trend is clear, AL methods improve upon random annotation methods (e.g., \bert{} compered to  \rbert), and DA approaches improve upon non-DA approaches (e.g., \rdt vs. \rbert). Importantly, we observe a complementary effect when combining AL and DA, with \tapad{} and \dt{} outperforming all other methods, with an average improvement of $2.0\%$ and $1.2\%$ (AUC) compared to \bert{} (their AL-based baseline) and an average improvement of $4.2\%$ and $4.6\%$ (AUC) compared to \rtapad{} and \rdt{} (their DA baselines). For SA, while UDA methods prove to be highly efficient (e.g., consider \rbert{} vs. \rdt), AL gains are inconsistent. On the one hand, \smoothing{} and \dropout{} improve \rbert{} by $0.6\%$ and $0.8\%$, respectively, on average. On the other hand, \rbert{} and \bert{} are equivalent. Nevertheless, it is \select{} and \dt{}, which combine AL and DA, that perform best, with an average improvements of $7.3\%$ and $3.6\%$, respectively, on top of the vanilla baseline, \rbert. 

Finally, we notice a limitation of the \select{} method, exhibiting dramatic performance drop in the NER setups. We analyze this behaviour in \S\ref{sec:ablation}.

\section{Conclusions}
\label{sec:conclusions}
This paper presents a new challenging learning setup where annotating data from the desired target domain is not feasible, and annotating data from available source domains is costly. 
We experimented with various prominent approaches within this setting, mainly from the active learning and domain adaptation algorithm families. We presented novel ways of combining the two, yielding the best performance.
We hope our findings will encourage further research on this setup, which we believe is common for real-world applications and should get more attention from the research community.

\section{Limitations}
\label{sec:limitations}

\paragraph{Future algorithmic solutions}
As mentioned in previous sections, we introduce a challenging setting for domain adaptation from scratch. Yet, we consider the algorithmic solutions proposed in this paper as an initial attempt. We focused on the intersection of active learning and domain adaptation and introduced novel algorithms that integrate both, while other approaches are also valid for the suggested setting. For instance, one can introduce self-training methods, apply low resource algorithms (zero-to-few shot learning), and perhaps consider combining these approaches with the ones we present in this work. We hope to inspire future work in the intersection of domain adaptation and training algorithms from the early production stages.

\paragraph{Inconsistent performance by the \select{} method}
We notice an interesting behavior of the \select{} algorithm, which performs well on SA (on average, outperforming all other models by more than $3.7\%$) while exhibiting a dramatic performance drop in NER setups. Such inconsistent behaviour across tasks, unless understood, may dissuade future implementation of this method. Thus, in \S\ref{sec:ablation}, we provide an extensive analysis of the \select{} algorithm behaviour across different tasks.

\newpage

\bibliography{acl2021}
\bibliographystyle{acl_natbib}

\newpage
\appendix


\begin{figure*}

\begin{minipage}{.33\textwidth}
\centering
\subfloat[\textbf{Colored domains}]{\label{fig:sa-tsne-all-domains}\includegraphics[scale=.15]{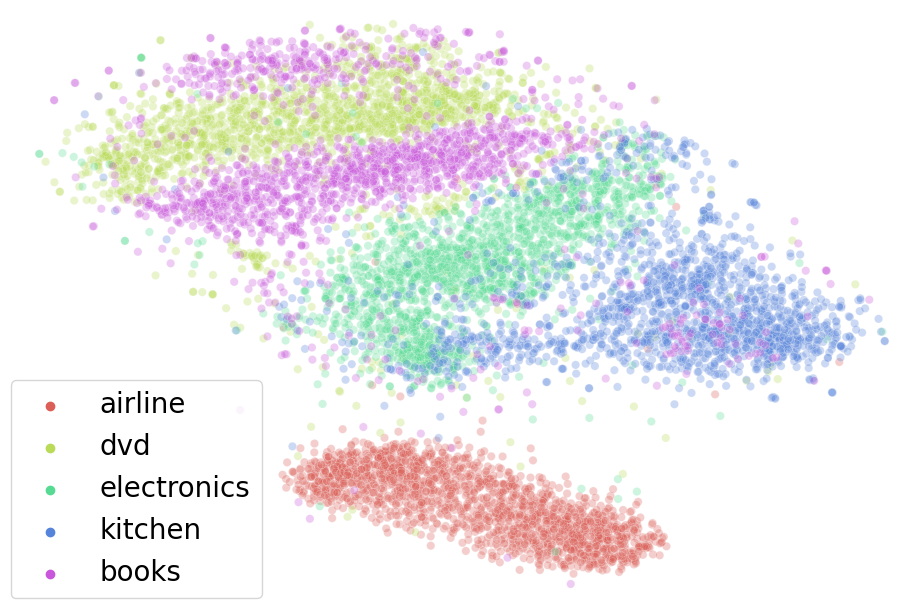}}
\end{minipage}
\begin{minipage}{.33\textwidth}
\centering
\subfloat[\textbf{\bert}]{\label{fig:sa-tsne-bert}\includegraphics[scale=.15]{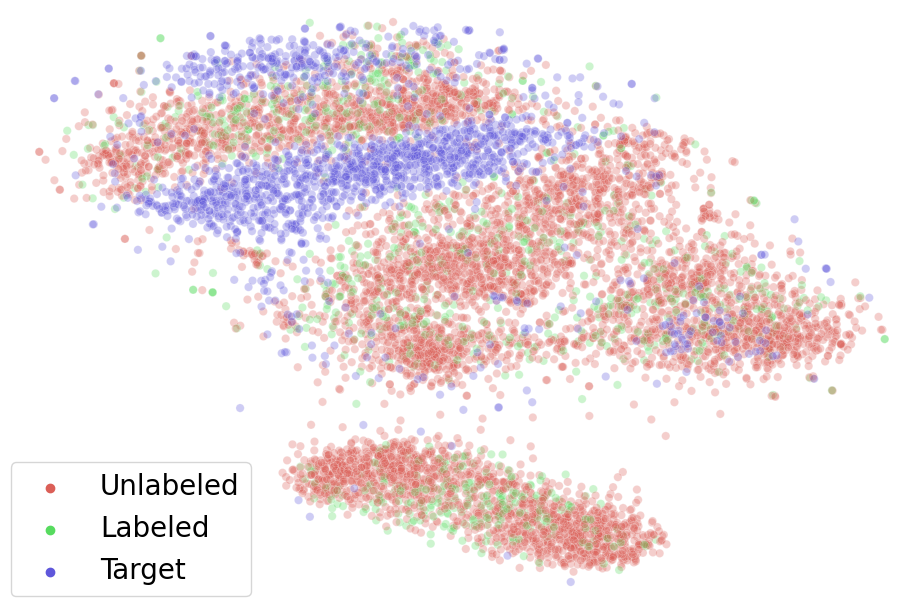}}
\end{minipage}
\begin{minipage}{.33\textwidth}
\centering
\subfloat[\textbf{\select}]{\label{fig:sa-tsne-selection}\includegraphics[scale=.15]{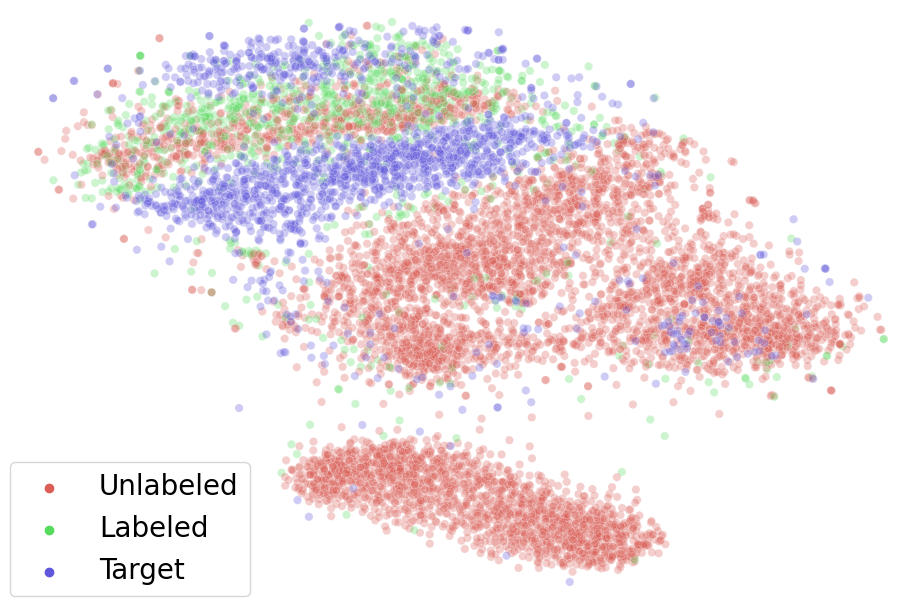}}
\end{minipage}

\begin{minipage}{.33\textwidth}
\centering
\subfloat[\textbf{Colored domains}]{\label{fig:ner-tsne-all-domains}\includegraphics[scale=.15]{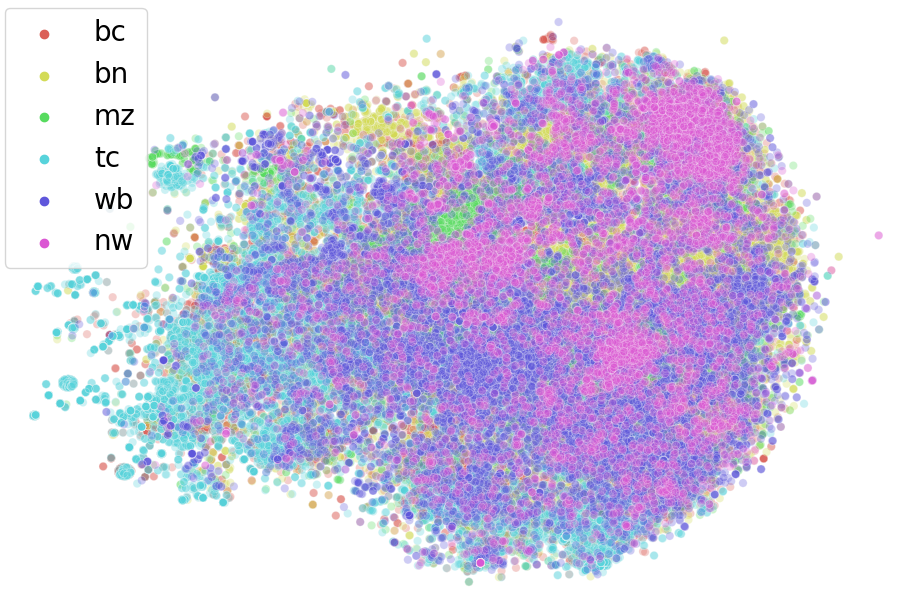}}
\end{minipage}
\begin{minipage}{.33\textwidth}
\centering
\subfloat[\textbf{\bert}]{\label{fig:ner-tsne-bert}\includegraphics[scale=.15]{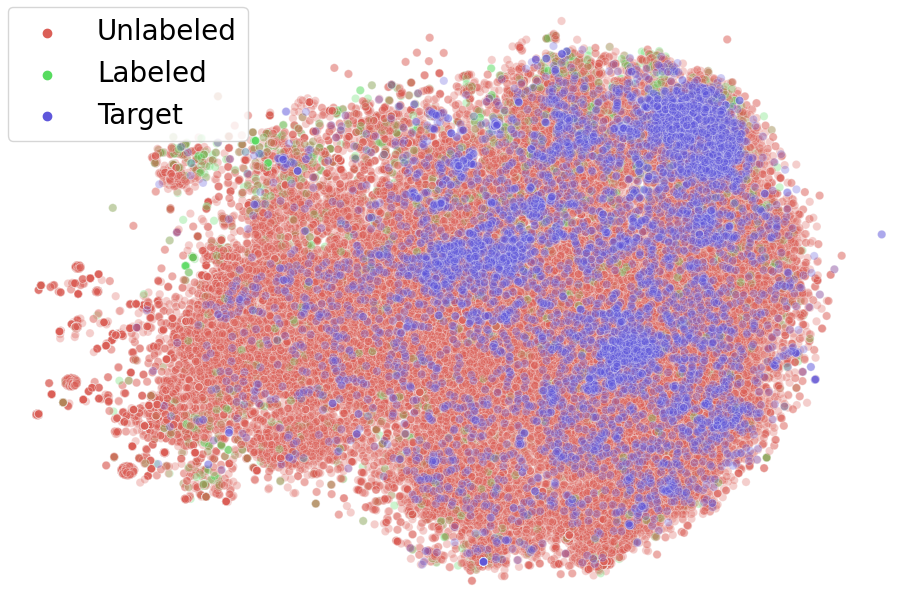}}
\end{minipage}
\begin{minipage}{.33\textwidth}
\centering
\subfloat[\textbf{\select}]{\label{fig:ner-tsne-selection}\includegraphics[scale=.15]{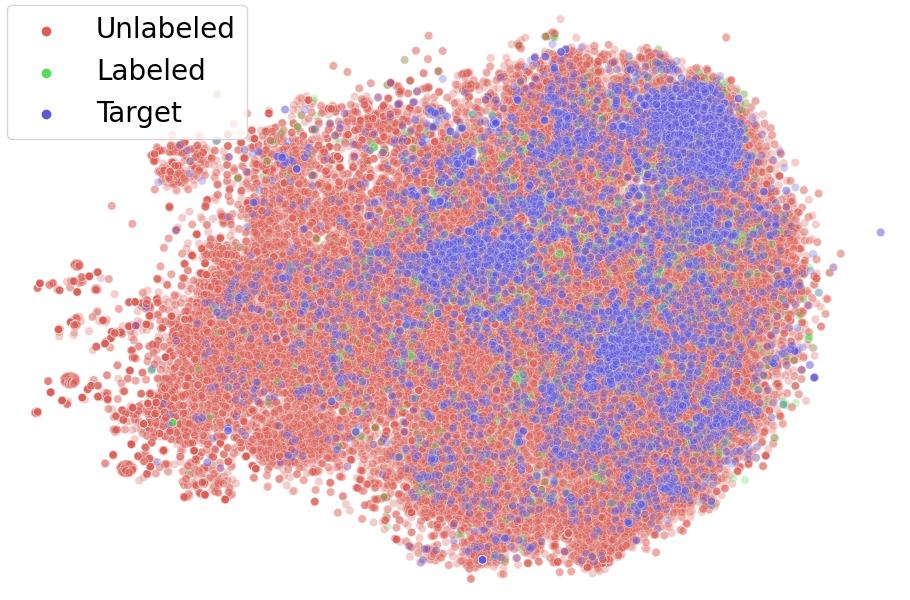}}
\end{minipage}

\caption{T-SNE visualization of the examples selected by \bert{} and \select. Top row graphs present results in the sentiment analysis task, where the target domain is \textit{Books}, while the graphs from the bottom row present results in NER, where the target domain is \textit{News (nw)}. }
\label{fig:tsne-ner-sentiment}
\end{figure*}

\section{Ablation Analysis}
\label{sec:ablation}

\paragraph{Analysing the performance drops of \select{} in NER}
To better understand the inconsistent behavior of the \select{} algorithm across tasks, we would like to analyze and understand the characteristics of  examples selected by the \select{} method for annotation. To this end, we project the entire corpus of both tasks to a 2D space by applying T-SNE to the sentence representations generated by BERT-base. We then choose a single target domain and plot three figures (per task): (1) visualization of  the domains, with dots of similar colors representing examples from the same domains (Figures~\ref{fig:sa-tsne-all-domains} and \ref{fig:ner-tsne-all-domains}); (2) visualization of examples selected for annotation by \bert{} (Figures~\ref{fig:sa-tsne-bert} and \ref{fig:ner-tsne-bert}); and (3) visualization of examples selected for annotation by \select{} (Figures~\ref{fig:sa-tsne-selection} and \ref{fig:ner-tsne-selection}).

Clearly, \select{} selects examples that are close to the target domain (examples nearby the purple cluster in Figure~\ref{fig:sa-tsne-selection} and \ref{fig:ner-tsne-selection}). In SA, examples from different domains are easily separated and identified. Accordingly, the algorithm mostly annotates examples from the closest available source domain. For instance, when the target domain is \textit{books}, \select{} annotates mostly from \textit{dvd}. As mentioned above, this leads to significant improvements.

In our NER experiments, the same behavior leads to a dramatic degradation in performance. We hypothesize that since NER is a multiclass task, with 38  labels, some of them are more frequent than others (unbalanced dataset), an algorithm that considers only the input representations may perform poorly. As demonstrated in Figures~\ref{fig:ner-tsne-all-domains} and \ref{fig:ner-tsne-selection}, \select{} chooses many examples from the \textit{wb} domain for annotation.\footnote{This is also the case for other target domains. As seen in Figure~\ref{fig:tsne-ner-sentiment}, \textit{wb} examples have representations that are close to examples from all other domains.} When taking a closer look into the statistics of this corpus, we notice that examples from \textit{wb} have fewer than 0.8 entities per example, on average, which is significantly lower compared to other domains. Finally, we rerun the \select{} algorithm, this time taking the \textit{wb} examples out of the annotation pool. As can be seen in Table~\ref{tab:no-web}, omitting the \textit{wb} domain from the annotation pool improves \select{} results, outperforming \rbert{} by $4.3\%$ on average. 

\begin{table}
\centering
\begin{adjustbox}{width=0.45\textwidth}

\begin{tabular}{ | l || c | c | c | c | c | c || c |}
\hline
  & \textbf{BC}  & \textbf{BN} & \textbf{MZ} & \textbf{NW} & \textbf{TC} & \textbf{WB} & \textbf{AVG}  \\
\hline

\textbf{\rbert} & $ 46.5 $ &  $ 48.5 $ & $ 42.1 $ & $ 36.2 $ & $ 32.5 $ & $ 30.7 $ & $ 39.4 $  \\
\textbf{\select} & $ 40.2 $ &  $ 48.2 $ & $ 40.9 $ & $ 43.1 $ & $ 12.8 $ & $ 28.4 $ & $ 35.6 $  \\
\textbf{\select (no wb)} & $ 46.6 $ &  $ 62.7 $ & $ 42.3 $ & $ 49.4 $ & $ 32.7 $ & $ 28.4 $ & $ 43.7 $  \\

\hline
\end{tabular}
\end{adjustbox}
\caption{AUC results for NER (using 2500 examples), in the condition where the \textit{wb} domain is discarded from the annotation pool.}
\label{tab:no-web}
\end{table}

\section{Implementation Details}
\label{sec:implementation}


\subsection{URLs of Code and Data}
\label{sub:urls}
\begin{itemize}
  \item \textbf{Our Code Repository} - our code will be published upon acceptance.
  \item \textbf{HuggingFace} \citep{wolf2020transformers} - code and pretrained weights for the T5 model and tokenizer: \url{https://huggingface.co/}
  \item \textbf{Bayesian Active Learning (BaaL) }\citep{baal} - a library which was used for the active learning process implementation: \url{https://github.com/ElementAI/baal}
  \item \textbf{Amazon Mechanical Turk}, a crowdsourcing platform: \url{https://www.mturk.com/}
\end{itemize}

\subsection{Data Statistics}
\label{sub:data-stats}
As we indicate in the main paper, we experiment with two different NLP tasks: sentiment analysis (SA) and named entity recognition (NER). For SA, we obtain five different domains (A, B, D, E, and K) and for NER we have six different domains (BC, BN, MZ, NW, TC, and WB). Table~\ref{tab:datasets} describes the data statistics of each domain. Specifically, we report the amount of unlabeled data and the total amount of labeled data (which is part of the annotation pool) that are available in each domain. 

Since we follow a leave-one-out experimental protocol (with multiple source domains and a single target domain), the annotation pool includes the entire available labeled data that is accumulated from all source domains. For instance, in SA, the size of the annotation pool is 8000, as we always have four source domains, with 2000 labeled examples in each. We also specify the amount of training data use for both tasks: for SA, we start with an initial seed of 50 examples (taken from the annotation pool), and end the annotation process with 250 labeled examples. For NER, we start with 500 examples and end with 2500 examples. 
Our test set includes only examples that originate from the target domain, thus we promise a non-overlapping train and test sets. 

Finally, we do not have access to unlabeled data that originates from the NER domains. Accordingly, we use the labeled NER data (without the labels) for UDA algorithms that require access to unlabeled data.

\subsection{Hyperparameter Different Choices}
\label{sub:hyperparameters}
Since we focus on extremely low resource learning scenarios, we assume no access to a large development set on which hyperparameters could be optimized. Accordingly, we base our choice of hyperparameters on previous work and practical considerations. We use a learning rate of $5 \cdot 10^{-5}$ and a maximum sequence length of $128$ which was chosen according to computational limitations. We tune the batch size value from the following options: $(8, 16, 32)$. For SA, we use a batch size of of $8$ and for NER we use a batch size of $16$.

\begin{table}
\centering
\begin{adjustbox}{width=0.48\textwidth}

\begin{tabular}{ | l || c | c | c  | c | }
\hline
\multicolumn{5}{|c|}{Sentiment Classification \citep{blitzer2007biographies, Nguyen2015airline}} \\
\hline
Domain & Unlabeled & Annotation Pool & Train  & Test  \\
\hline
Airline (A) & 39454 & 2000 & 50$\rightarrow$250  & 2000 \\
\hline
Books (B) & 6001 & 2000 & 50$\rightarrow$250 & 2000 \\
\hline
DVDs (D) & 34742 & 2000 & 50$\rightarrow$250 & 2000 \\
\hline
Electronics (E) & 13154 & 2000 & 50$\rightarrow$250 & 2000 \\
\hline
Kitchen (K) & 16786 & 2000 & 50$\rightarrow$250 & 2000 \\

\hline
\multicolumn{5}{|c|}{Named Entity Recognition \citep{ontonotes}} \\
\hline
Domain & Unlabeled & Annotation Pool & Train & Test  \\
\hline
Broadcast Conversations (BC) & - & 11880 & 500$\rightarrow$2500 & 7515  \\
\hline
Broadcast News (BN) & - & 10684 & 500$\rightarrow$2500 &  8370 \\
\hline
Magazine (MZ) & - & 6774 & 500$\rightarrow$2500 &  8946 \\
\hline
Newswire (NW) & - & 34970 & 500$\rightarrow$2500 &  7400 \\
\hline
Telephone Conversation (TC) & - & 12892 & 500$\rightarrow$2500 &  8360 \\
\hline
Web (WB) & - & 15642 & 500$\rightarrow$2500 &  8040 \\
\hline
\end{tabular}
\end{adjustbox}
\caption{Number of available samples in each domain. Right arrows ($\rightarrow$) mark the amount of labeled data used as the initial AL seed (at the left-hand side of the arrow) and the size of the training data at the end of AL process (at the right-hand side of the arrow).}
\label{tab:datasets}
\end{table}

\subsection{Computing Infrastructure and Runtime}
\label{sub:compute}
All experiments were performed on either one or
two Nvidia GeForce GTX 1080 Ti GPUs, with two
cores, 11 GB GPU memory per core, 6 CPU cores
and 62.7 GB RAM.

In sentiment analysis, we measured an average of 5 minutes for running a single training experiment with \rbert, \bert, \smoothing, \rtapad, and \tapad. For \rdann{} and \dann{} we measured an average of 10 minutes, for \dropout{} we measured 15 minutes, and for \rdt{} and \dt{} it takes 10 minutes to finetune the model on the unlabeled data.
In NER, we measured an average of 10 minutes for running a single training experiment with \rbert, \bert, \smoothing, \rtapad, and \tapad. For \rdann{} and \dann{} we measured an average of 15 minutes, for \dropout{} we measured 15 minutes, and for \rdt{} and \dt{} it takes 13 minutes to finetune the model on the unlabeled data.

\end{document}